\documentclass[11pt,twocolumn]{article}
\usepackage{multirow}
{}
{}
{}
\usepackage{xspace}
\usepackage{subcaption}
\usepackage{graphicx} 
\usepackage{colortbl}
\usepackage[dvipsnames]{xcolor}
\definecolor{green}{rgb}{0.1,0.1,0.1}
\newcommand{\firstplus}{\cellcolor{blue}\textcolor{white}{$+++$}}  
\newcommand{\secondplus}{\cellcolor{teal}\textcolor{white}{$++$}}  
\newcommand{\thirdplus}{\cellcolor{cyan}\textcolor{white}{$+$}}  
\newcommand{\neutral}{\cellcolor{white}$=$}  
\newcommand{\firstminus}{\cellcolor{pink}\textcolor{white}{$-$}}  
\newcommand{\secondminus}{\cellcolor{magenta}\textcolor{white}{$--$}}  
\newcommand{\thirdminus}{\cellcolor{red}\textcolor{white}{$---$}}  
\usepackage{url}
\usepackage[export]{adjustbox}
\usepackage{authblk}
\usepackage{hyperref}

\begin{document}

\title{Performance of computer vision algorithms for fine-grained classification using crowdsourced insect images}

\author[$\dagger$]{Rita Pucci, email \href{mailto:rita.pucci@naturalis.nl}{rita.pucci@naturalis.nl} }
\author[$\dagger$]{Vincent J. Kalkman}
\author[$\star$,$\dagger$]{Dan Stowell}

\affil[$\dagger$]{Naturalis Biodiversity Center, Leiden(NL)}
\affil[$\star$]{Department of Cognitive Science and AI, Tilburg University(NL)}

\def\myeffn{\texttt{EffNet}\@\xspace}
\def\myincpt{\texttt{Incpt}\@\xspace}
\def\myresNet{\texttt{ResNet}\@\xspace}
\def\myttv{\texttt{T2TViT}\@\xspace}
\def\myconvit{\texttt{ConViT}\@\xspace}
\def\mydeit{\texttt{DeiT}\@\xspace}
\def\myvitae{\texttt{ViTAE}\@\xspace}
\def\myvitdeffn{\texttt{ViTdEfN}\@\xspace}
\def\myvitdvitae{\texttt{ViTdVAE}\@\xspace}
\def\myAE{\texttt{AE}\@\xspace}
\maketitle
\section{Abstract}
With fine-grained classification, we identify unique characteristics to distinguish among classes of the same super-class. We are focusing on species recognition in Insecta, as they are critical for biodiversity monitoring and at the base of many ecosystems. With citizen science campaigns, billions of images are collected in the wild. Once these are labelled, experts can use them to create distribution maps. However, the labelling process is time-consuming, which is where computer vision comes in. The field of computer vision offers a wide range of algorithms, each with its strengths and weaknesses; how do we identify the algorithm that is in line with our application? To answer this question, we provide a full and detailed evaluation of nine algorithms among deep convolutional networks (CNN), vision transformers (ViT), and locality-based vision transformers (LBVT) on 4 different aspects: classification performance, embedding quality, computational cost, and gradient activity. We offer insights that we haven't yet had in this domain proving to which extent these algorithms solve the fine-grained tasks in Insecta. We found that the ViT performs the best on inference speed and computational cost while the LBVT outperforms the others on performance and embedding quality; the CNN provide a trade-off among the metrics.

\section{Introduction}\label{sec1}
\label{sec:intro}
The fine-grained classification task is aimed at distinguishing between different classes that belong to the same super-class. In the field of biodiversity monitoring, species are considered as classes, and the super-class is the order, e.g.~\cite{chang2017fine}. Taxonomists, who are domain experts, define the species based on their morphology or molecular data. Since the species within an order are closely related and share similar characteristics, such as colours and traits, fine-grained classification tasks in this field are particularly challenging. In this paper, we will consider the taxonomic class of Insecta and in particular the orders of Coleoptera and Odonata focusing on the European species. Coleoptera and Odonata are two of the oldest insect orders and play important roles in our environment. However, they are still not fully understood due to the complexity involved in identifying the different species. Coleoptera can be further divided into four suborders: Archostemata, Myxophaga, Adephaga, and Polyphaga, with over 130,000 species present in Europe alone~\cite{audisio2015fauna}. The order Odonata can be divided into two suborders: Epiprocta and Zygoptera, with over 200 species present in Europe alone. 
\begin{figure}[!b]
\centering{\includegraphics[width=\columnwidth ]{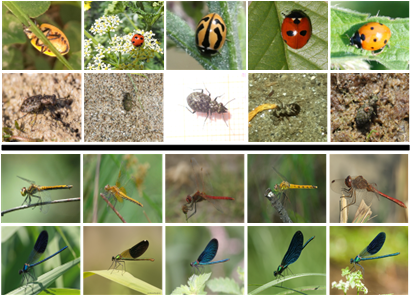}}
\caption{Images in each group belong to the same order, images in each row belong to the same genus, and each image represents a unique species from the Coleoptera and Odonata.\label{fig:one}}
\end{figure}
Many species within both orders have similar physical characteristics, making it difficult to differentiate between them. 
Fig.~\ref{fig:one} shows sample images for Observation.org dataset~\cite{Obs} from the two classes which exhibit low inter-species variations when the species are from the same genus and large intra-species variations otherwise. Additionally, in Coleoptera, many species are small, and their distinguishing features are often hard to discern. Other factors that make identification challenging include within-species variability arising from differences in life stages, sexes, and regional or seasonal variations. 

The ability to identify the insects that inhabit ecosystems is one of the main steps to understanding them. Despite its significance, the fine-grained task in biodiversity has posed two key challenges: 1) Inter-class variances are often extremely subtle, thus requiring highly discriminative representation for effective classification; 2) As the rarity of a species increases, there are fewer training samples per category, impeding the performance of large-data favoured methods. The conventional identification technique is to cross-validate the image with the regional field guides, online sources, and field experts. The majority of the images of these insects are collected by citizen scientists. The use of tools such as Obsidentify~\cite{Obsidentify} helps them in collecting images which are going to be part of the datasets studied by expert taxonomists and available on data platforms such as GBIF~\cite{GBIF}. This ongoing work is crucial to improve the knowledge of the current state of biodiversity. Now, billions of images are available for insects and conventional identification techniques cannot stand alone since they are highly time-consuming and unaffordable for the common person. There is increasing interest in the investigation of new deep learning fine-grained methods for biodiversity monitoring. Early and fast identification techniques are crucial and the fast-developing of deep learning technologies in computer vision have shown impressive solutions to many real-world problems such as animal identification~\cite{tuia2022perspectives}. At the state of the art, the convolutional neural network (CNN) for computer vision is an algorithm based on an inductive bias of locality and shift-invariance. These two main features make CNN a highly effective deep learning algorithm in image classification. Many variations of the CNN algorithm are available addressing different limitations and proposing new and advanced structures. We have seen an increased interest in the application of transformers for the same tasks to which CNN was historically devoted. Vision transformer (ViT)~\cite{dosovitskiy2010image} enables multi-head self-attention to capture long-range dependencies within an image and thus can extract diverse feature patterns for discriminative classification. Unfortunately, ViT is data-hungry and the lack of training data may impede its application in fine-grained tasks. As for CNN also with ViT, at the state of the art, we can find different approaches to overcome the limits such as knowledge distillation~\cite{hinton2015distilling}. Recently, a merged group of algorithms is taking space in this challenge: the locality-based vision transformers (LBVT). As the name suggests, these algorithms are based on vision transformers and then improved with modules of a structure composed of convolutional layers~\cite{liu2022cross}. The obtained algorithms benefit from the inductive bias of locality and shift invariance from the convolutional layers while being able to capture long-range dependencies with self-attention modules. On the other hand, the structure and order of such layers require a deeper understanding.

With their pros and cons, these algorithms are all good candidates for fine-grained tasks for insect images but to what extent? Is the classification performance a sufficient metric to evaluate these algorithms? 

In this paper, 
\begin{itemize}
    \item We take into consideration these three groups of deep learning models and we delve into their behaviour when applied to fine-grained tasks in the biodiversity monitoring domain.
    \item We evaluate each model on four aspects each of which will put the algorithms on different prospects giving us an intuition on what to expect from the model. These are:
    \begin{itemize}
        \item the classification performance.
        \item the embedding quality.
        \item the computational cost.
        \item the gradient activity.
    \end{itemize}
    \item We present an overall and a per-species classification performance analysis to observe their behaviour with the long tail distribution of species and in particular with rare species. 
    \item Based on the compressed input representation, each of these models creates an embedding space. We evaluate the quality of the embedding space based on its ability to capture and represent the underlying structure and relationships in the data. 
    \item We consider one critical key point which influences the use of an algorithm, which is the computational costs as the resources required, the computational time, and the computational complexity. 
    \item To answer the question of which part of the image affects the prediction of the model, we will analyse the gradient activation with GradCam.
\end{itemize}
For this study, we select algorithms at the state of the art which obtain the best performance or are commonly used in image classification: Inception\_v3~\cite{szegedy2017inception}(\myincpt), EfficientNet\_v2~\cite{tan2104efficientnetv2}(\myeffn), ResNet 50~\cite{reil2022resnet}(\myresNet) for the convolutional neural network (CNN); T2TViT\_14~\cite{yuan2023t2tvit}(\myttv), and ViT trained in knowledge distillation for Vision Transformers (ViT); ConViT~\cite{d2021convit}(\myconvit), and ViTAEv2~\cite{zhang2023vitaev2}(\myvitae) for the Locality-based Vision Transformers (LBVT). 

For training and validation, we consider datasets collected by citizen science and stored in Observation.org~\cite{Obs}. For the classification performance, we evaluate the models on Artportalen~\cite{art} limited to Odonata and Coleoptera from Europe, which are collected from different communities of citizen science than training. All four analyses present the results to address the fine-grained task at the species level. 

Fine-grained accuracy for biodiversity monitoring is a difficult task, which is why our comprehensive evaluation of 9 different computer vision models based on 4 distinct aspects provides a unique contribution to the field. Our evaluation offers a detailed assessment of the competing paradigms for neural network architectures, which is something that has been missing until now. We think that by leveraging our results, we can advance the development of more effective and efficient classification techniques.

\section{Related work}\label{sec2}
Insecta are the most diverse taxonomic class of animals on earth, but their small size and high diversity have always made them challenging to study. Extensive work has been done to monitor different species in different orders e.g. Lepidoptera, Coleoptera, Odonata, Orthoptera, and Hymenoptera. Monitoring activities determine which insects are at risk, how insect populations fluctuate in natural areas, and which management actions are most beneficial to the ecosystems~\cite{theivaprakasham2021identification, chang2017fine, ding2016automatic,lim2017performance,xia2018insect,dembski2019bees,theivaprakasham2022odonata}. An important resource is the data collected by citizen sciences around the earth, but that implies an enormous amount of data to classify. In this context, the application of deep learning algorithms, such as convolutional neural networks (CNN), has seen increased popularity for the ability of automated feature extraction and a high accuracy rate in fine-grained classification. 
CNN is now popularly used for insect identification and presents a wide range of models applied to classify species in different case studies. Works propose species classification in Lepidoptera order~\cite{theivaprakasham2021identification, chang2017fine, ding2016automatic} which reach high performance in accuracy. Customized models are proposed for generic species from different orders in the class Insecta~\cite{lim2017performance,xia2018insect}, also specifically to classify bees in real time~\cite{dembski2019bees}, Orthoptera for mobile application~\cite{chudzik2020mobile}, and Odonata~\cite{theivaprakasham2022odonata}. 
Even if we have a prolific application of dedicated CNN in fine-grained tasks for insects, we demonstrate in the paper that these models have limitations on the identification of rare species, extraction of compressed input representation and memory efficiency resulting in enormous limitations in practical applications. It is still an open challenge that requires investigation. Moreover, we do not observe equal interest in the application of transformer-based models in this task. An interesting comparison between very simple CNN and transformer-based algorithms for fine-grained tasks among species of different kingdoms identifies the ViT model as outperforming the CNN-based models~\cite{peng2022cnn}. A customised transformer model is proposed for insect pest recognition highlighting the need to integrate some of the CNN features into the transformer structure making the model focus more on global coarse-grained information rather than local fine-grained information~\cite{wang2023aa}. 

Though a vast amount of work has been done in the domain of insect identification, we have not found any published research on a comparative evaluation of algorithms from three of the main groups of deep neural networks for fine-grained identification in biodiversity monitoring. Furthermore, there is no experimentation on the most modern models from computer vision for this task. 

\section{Methods}\label{sec3}
\subsection{Deep Learning Models}
Among all the groups of models available in computer vision, we select three groups of deep neural network models that are widely used at the state of the art for classification in computer vision for ecology. For each of these, we select the latest algorithms with the best performance on ImageNet1K~\cite{russakovsky2015imagenet} and some of the most used algorithms for image classification. For a fair comparison, none of these models are specialised for fine-grained tasks.

\paragraph{\textbf{Convolutional Neural Networks}} 
In the group of CNN, we consider models that are mainly based on the convolutional layers and fully connected layers~\cite{mao2021theory}. We are interested in models that are competitive with ViT in inference speed, and model size. We choose Inception\_v3~\cite{szegedy2016rethinking}, EfficientNet\_ v2\_ medium~\cite{tan2104efficientnetv2}, and ResNet50~\cite{he2016deep}. In the paper, we refer to these models respectively as: \myincpt, \myeffn, and \myresNet. The \myincpt is the result of dealing with the trade-off between performance in classification tasks and computational cost. The structure is thought to scale up in a way that aims to utilise factorised convolutions to reduce the computational bottleneck due to fast and extreme compression of feature maps (convolution with kernels bigger than $3x3$). This idea makes the network wider instead of deeper in favour of an efficient computation~\cite{szegedy2016rethinking}. The \myeffn has the structure and connections optimised for speed, based on floating point operations per second (FLOPs), and for parameter efficiency. In particular, \myeffn consists of convolutional-based layers~\cite{gupta2019efficientnet, gupta2020accelerator} designed to better utilise mobile or server accelerators.  Both these models represent good competitors to the transformer-based models. The \myresNet is not a model of the same complexity as the previous two but it is one of the most used models as the backbone or the main model in ecology for classification~\cite{zhang2021tree, chen2021new, cao2020improved, reil2022resnet}. The CNN models are naturally equipped with intrinsic inductive bias, shift-invariance, and hierarchical structure to extract multi-scale features and locality. These are proper advantages in extracting representative features which are used to identify the species in the images. Even if CNN models are commonly used as the backbone of many image classification models at the state of the art, they are not well suited to model long-range dependency due to their structure focused on extracting local features from low level to high level progressively. This can affect the performance in the fine-grained tasks: these models are less inclined to identify relations among details of the subject. The details and their relationship are typically the characteristics used by taxonomists to distinguish species.

\paragraph{\textbf{Vision Transformers}} 
Models based only on attention~\cite{vaswani2017attention} are here referred to as fully-transformer models. The Vision Transformer (ViT)~\cite{dosovitskiy2010image} is the first fully-transformer model applied for image classification demonstrating that transformers are promising for vision tasks. ViT is based on the self-attention mechanism which allows the model to capture global contextual information, enabling it to learn long-range dependencies and relationships between image tokens (patches). The self-attention mechanism weighs the importance of different tokens in the sequence when processing the input data. In this paper, we consider for comparison the recent evolution of ViT, the Token-2-Token ViT 14~\cite{yuan2023t2tvit} (\myttv) which uses a progressive tokenization module to aggregate neighbouring tokens into one token. In the first layer, a token is a patch of the image, while in the intermediate layers, a token is a patch of the feature maps. The model can extract local information reducing the length of the token iteratively. This architecture reduces the data hunger and boosts the performance relative to the vanilla ViT. In addition to \myttv, we also consider the vanilla ViT\_small\_patch16~\cite{touvron2021training} model trained via knowledge distillation~\cite{hinton2015distilling, abbasi2020modeling}. \textbf{Knowledge distillation} is a model compression method in which a small model (the student) is trained to mimic a pre-trained, larger model (the teacher). In this paper, we follow the cross-architecture knowledge distillation~\cite{liu2022cross}, and we explore the use of homologous (both CNNs or Transformers) or not homologous architectures. We propose this technique in two flavours which consist of distillation from a CNN model, in particular \myeffn, ResNet like-model, and distillation from an LBVT model, \myvitae. In both, the ViT model is used as the student model. In the paper, we refer to these models respectively as: \myvitdeffn, \myvitdvitae. For these models, the teachers are trained on the training split of the case studies considered in the paper. In particular, the teachers are the \myeffn and \myvitae trained for this paper. Finally, we consider the data-efficient image transformers~\cite{touvron2021training} \mydeit. In this case, the teacher is a CNN model pre-trained on ImageNet1K. 

\paragraph{\textbf{Locality-based vision transformer}} Finally we consider Locality-based vision transformer models (LBVT), which create a collaboration between the convolutional and the transformer layers. With this intent, the CNN structures are included in the vision transformers since the convolution kernels help the model capture the local information. As such, adding locality from CNN improves the data efficiency of vision transformers, resulting in a better performance on a small dataset~\cite{ruan2022vision}. In this paper, we consider the Convolution-like ViT basic ~\cite{d2021convit}, and ViT Advanced by Exploring inductive bias version2 basic~\cite{zhang2023vitaev2}. We refer to these models as \myconvit, and \myvitae. The \myconvit model includes Gated Positional Self-attention(GPSA), which can be initialized as a convolutional layer~\cite{cordonnier2019relationship} for capturing the local information at the beginning of the training stage. As such, \myconvit can utilise the advantages of the soft inductive bias without being limited to CNN. GPSA allows ViT to be the same as CNN to improve the data efficiency on small datasets. The \myvitae implements inductive bias and the scale-invariance properties into a transformer architecture. To obtain such a result, the algorithm exploits multiple parallel convolutional layers to create the scale invariance and inductive bias, and the transformer layers to create long-range dependencies among the extracted features.

\begin{figure}
\centering
  \includegraphics[width=.9\linewidth]{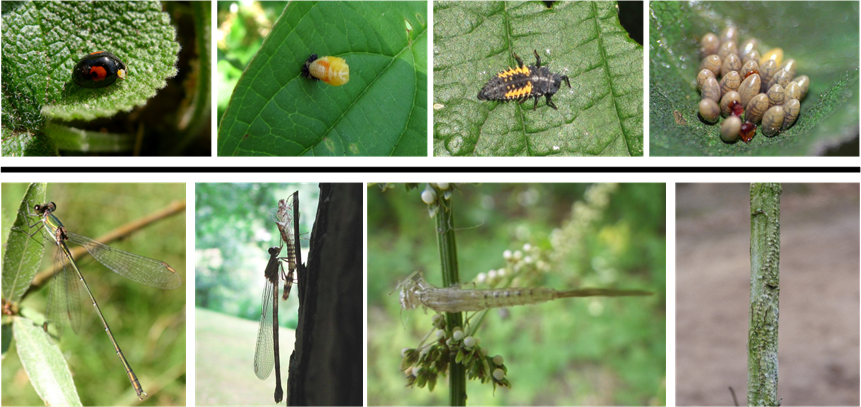}
  \caption{Metamorphism of Coleoptera (top row) and Odonata (bottom row) at different life stages.}
  \label{fig:morph}
\end{figure}
\subsection{Datasets}
\label{dataset}
The Coleoptera and Odonata datasets used in this paper are available on  Observation.org~\cite{Obs}, the largest nature platform in the Netherlands for nature observation. Each of these datasets is split into train and test datasets, by using only the test set we wish to evaluate fine-grained classification in highly-unbalanced natural datasets, with as few as 2 images per species in many cases. The models are trained only on the data available on the training set. The level of fine-grained tasks considered in this paper is the species level. In the quantitative analysis Sec.~\ref{Sec:class_metrics}, we evaluate the models on two datasets: the test split of the Observation.org datasets, and the Odonata and Coleoptera from Artportalen.se~\cite{art} -the Swedish nature conservation portal. With the test on the Artportalen.se datasets, we evaluate the generalisation ability of the trained models on a test set taken from an entirely different data source.
 In both orders, we consider only European species. The images are collected with mobile phones by citizen scientists. It is worth noting that many species of Coleoptera and all species of Odonata present sexual dimorphism and metamorphism due to different stages in life. Fig.~\ref{fig:morph} shows the challenges introduced with the dimorphism. For the same species, the animal can appear in completely different forms. With species, we refer to the taxonomical full name of the species which consists of the order (e.g. Odonata), the infraorder (e.g. Anisoptera), the family (e.g. Aeshnidae), the genus (e.g. \textit{Aeshna}), and the species (e.g. \textit{affinis}). 

\paragraph{\textbf{Coleoptera\_Obs}} The dataset, from Observation.org, consists of 849,296 images over 3,087 species. We split the dataset in train and test with the ratio of 80:20 samples per species (674,441:174,855 samples). The dataset is unbalanced with a minimum of 2 samples and a maximum of 11,523 samples per species. The dataset consists of species from 122 families, and we have samples from the Polyphaga and Adephaga suborders and 1,344 genera with a total of 3,087 species. In the dataset, there are samples of thirteen morphs: imago, imago brachypterous, imago macropterous, imago micropterous, unknown, gall, exuviae, deviant, larva/nymph, mine, egg, pupa, queen; and three sexes: male, female, and unknown. 

\paragraph{\textbf{Odonata\_Obs}} The Odonata dataset from Observation.org contains 628,189 images from 235 wild Odonata species. The ratio of train data and test data is roughly 80:20 per species (502,467:125,722 samples). The dataset is unbalanced with a minimum of 2 samples and a maximum of 19,754 samples per species. 
The dataset consists of species from both Epiprocta, in particular from Anisoptera, and Zygoptera infraorders. For the Anisoptera infraorder, we have samples from six families for a total of 153 species, and for the Zygoptera infraorder, we have five families for a total of 82 species, we name this dataset with Odonata\_Obs. For this dataset, information is available about morph and sex for further observation in the results. The dataset consists of samples of eight morphs: imago, unknown, fresh imago, exuviae, deviant, larva/nymph, prolarva, and egg; and three sexes: male, female, and unknown.

\paragraph{\textbf{Coleoptera\_Art}} The Artportalen~\cite{Artportalen_Coleoptera} consists of 3,426 species of Coleoptera. Among them, 1,574 are used to validate the models and are available in the train split of Coleoptera\_Obs. The dataset is unbalanced with a total of 118,464 samples. There are more than 400 species with less than 10 samples each and less than 30 species with more than 500 samples each. 

\paragraph{\textbf{Odonata\_Art}} The Artportalen~\cite{Artportalen_Odonata} has 73 species from both Anisoptera and Zygoptera infraorders of which 69 species are available in the train split of Observation.org. The dataset consists of 55,680 samples and it is unbalanced with 12 species with less than 100 samples and 20 species with more than 1,000 samples.

\subsection{Experimental configuration}\label{sec4}
\subsubsection{Preparation of the models}
All the models considered in this paper are pre-trained on ImageNet1k with exceptions for the models which serve as teachers in \myvitdeffn and \myvitdvitae. We used the checkpoints available online for the \myvitae~\cite{zhang2023vitaev2} and \myttv~\cite{yuan2023t2tvit}. For the pre-trained initialization of \myeffn, \myincpt, \myresNet, \myconvit, \mydeit, and the ViT model in \myvitdeffn and \myvitdvitae, we used the checkpoints available on PyTorch Image Models (timm)~\cite{rw2019timm}.  All the models are modified in the head layer, which is replaced with a linear layer for the number of classes to be identified. The models are fine-tuned on the dataset described in Sec.\ref{dataset}. We perform a complete fine-tuning, which means all the parameters of the architectures are updated during the training phase. All the experiments are executed on NVIDIA A40 GPU.
We trained the model for a maximum of 310 epochs. We apply early stopping regularisation based on training loss to avoid overfitting.
We used a batch size of 32 samples, $5 \times 10^{-4}$ as the learning rate, 0.065 weight decay, with  AdamW~\cite{kingma2014adam} as the optimiser with cosine learning rate decay~\cite{loshchilov2016sgdr}, and 10 warm-up epochs. Due to limited resources, we analysed only one run.
\subsubsection{Augmentation and data preparation}
\label{sec:pre-processing}
For a fair comparison, we implement the same training scheme for all three models. We set the image size as $224\times 224$ and apply augmentation methods: mixup~\cite{zhang2017mixup}, and cutmix~\cite{yun2019cutmix} for all the models. We do not apply any balancing process and we evaluate the models on the test set available for the datasets.

\subsection{Classification metrics}\label{Sec:class_metrics}
The performance in the classification of the models is evaluated with the entire test set of the dataset and with each species separately. The evaluation of the models on the dataset is presented for the test set of Observation.org and Artportalen datasets. We take into consideration two metrics to evaluate the models: average batch accuracy (avgACC), and F1score. We consider avgACC top-1, hereafter named top-1, the model prediction with the highest probability must be exactly the expected answer; and avgACC top-5, hereafter named top-5, which considers any of our models' top-5 highest probability answers to match with the expected answer. The F1score evaluate the weighted average mean of Precision and Recall. To evaluate the performance of our models, we analyse their accuracy when applied to each species individually. We calculate the average accuracy achieved on all samples of a given species. Our analysis includes results for both rare and common species and also identifies the number of species that were not recognised by the models. Additionally, we examine the confusion matrices of each model to understand the nature of any misclassifications.

\subsection{Embedding metrics}\label{Sec:embed_metrics}
Clustering refers to the process of partitioning a dataset into different groups, called clusters. The data points in each cluster share similar characteristics or properties. Why is it important in this context? We consider the clusters in the embedding space provided by each model. The embedding space is a relatively low-dimensional space into which it is translated an high-dimensional vectors. For each input, the model compresses the information in an embedding which populates the embedding space. We evaluate if the embedding of each of the samples in the test set is organised in distinctive clusters. As described in Sec.~\ref{dataset}, all our data are annotated, and the annotation is made with taxonomic names, where both the genus and species are available. We consider the correspondence between clusters obtained with genus and species labels to evaluate the models.
We present, in~\ref{Sec:embed_perm}, a quantitative evaluation of the clusters formed in the embedding space. With this intent, we use the Silhouette Score which aids in the assessment of clustering performance. For each data point, it considers the average distance of the point to all other data points in the same cluster (intra-cluster distance), and the average distance of the point to all data points in the nearest cluster (inter-cluster distance). The values presented in the section are the overall Silhouette Score computed as the average value among all the points. We then present a visual representation of the embedding distribution for each model obtained with Uniform Manifold Approximation and Projection (UMAP)~\cite{mcinnes2018umap}. The UMAP is a dimension-reduction technique that can be used for visualisation. UMAP reduces the data after learning the manifold. It is based on parameters such as \textit{n\_neighbors} - the size of the local neighbourhood used in learning the manifold, \textit{min\_dist} - how tight the points are packed together, and \textit{n\_components} - dimensionality of the reduced dimension space (2D/3D). We set these parameters based on our empirical analysis with values: \textit{n\_neighbors} = 50, \textit{min\_dist} = 0.5, and \textit{n\_components} = 2. 

\subsection{Computational cost metrics}\label{Sec:cost_metrics}
The computational cost is the measure of the amount of resources a neural network uses in training or inference. This analysis is important to evaluate the feasibility of the models. We will analyse metrics which refer to the performance of the models in terms of the time demand, the computing power, and the memory space needed by each model. 
The models are here evaluated based on the structural information which is invariant with the case studies considered, and the time and memory demand which is closely related to that. In this paper, we evaluate the number of layers, FLOPS, inference and training times, and the number of parameters.
The number of layers, in convolutional-based models, expresses the capacity of the model to compress the features while extracting them. The floating point operations per second (FLOPS) is a measure of the computational complexity of deep learning models. It describes how many operations are required to run a single instance of a given model. We report the FLOPS results presented in the papers at the state-of-the-art (venue). The inference time is here computed as the time needed for a model to provide a prediction of a batch of one image. We compute the inference time as the mean out of three predictions. We load the fine-tuned model, we load an image and we apply the transformations required (resize, and cast in a tensor) and we execute the model in evaluation modality, by using the utility of PyTorch. The inference time is related only to the computation of the prediction. We consider the number of parameters that are trained in the structure, these parameters define the memory required by the model to be loaded. The number of trainable parameters affects the training time, which is here considered as a metric to evaluate the computational cost of the models. The training time is computed as the time needed to compute the prediction, the loss, make the backward step to accumulate the gradient for each parameter, and the optimisation step to update all the parameters based on the current gradients. The inference and the training time are computed as the mean time required on three executions on our hardware.

\subsection{Gradient activity metrics}\label{Sec:gradCam_metrics}
To explain how the model behaves with the input provided, we use the gradient-weighted class activation mapping (Grad-CAM)~\cite{jacobgilpytorchcam} technique. Grad-CAM utilises the gradients of the classification score concerning the final layers of each model, to identify the parts of an input image that most impact the classification score. The places where the gradient is large are where the final score depends most strongly on the data. 
With models in the CNN group, the Grad-CAM uses the feature maps produced by the batch normalisation layer in the last convolutional layer of a CNN. With models in the ViT and LBVT groups, we use the output of the normalisation layer before the attention block. This output is a tensor $<B,W,H>$, where $B$ is the batch dimension, $W$ consist of the class token, and the patches that make up the image, and finally $H$ are the channels. To reshape the tensor to the 2D spatial images, we use a reshape\_transform function as defined in~\cite{jacobgilpytorchcam}. 
\begin{table*}
    \centering
    \begin{tabular}{|c|c|c|c|c|c|c|c|}\hline
 \multicolumn{2}{|c|}{NN}& \multicolumn{3}{|c|}{Coleoptera\_Obs}& \multicolumn{3}{|c|}{Odonata\_Obs}\\\hline \hline 
         Model&  Version &  Top 1&  Top 5&  F1  &  Top 1&  Top 5& F1 \\ \hline 
         \myincpt&  base\_v3 &  $86.10\%$ &  $95.60\%$ &  $85.75\%$ &  $89.90\%$ &  $97.40\%$ & $88.54\%$ \\ \hline 
         \myeffn&  m\_v2 &  $88.00\%$ &  $96.70\%$ &  $87.78\%$ &  $92.60\%$ &  $98.50\%$ & $\textbf{94.30\%}$ \\ \hline 
         \myresNet&  50 &  $86.70\%$ &  $95.90\%$ &  $86.40\%$ &  $90.80\%$ &  $97.80\%$ & $90.41\%$ \\ \hline 
         \myttv&  14 &  $88.10\%$ &  $96.70\%$ &  $87.97\%$ &  $91.50\%$ &  $98.40\%$ & $93.65\%$ \\ \hline 
         \mydeit&  B\_p\_16&  $84.10\%$ &  $95.10\%$ &  $83.90\%$ &  $89.70\%$ &  $97.70\%$ & $89.32\%$ \\ \hline 
         \myvitdvitae&  S\_p\_16&  $ 86.60\%$&  $96.10\%$&  $86.11\%$&  $90.90\%$&  $98.30\%$& $90.33\%$\\ \hline 
         \myvitdeffn&  S\_p\_16&  $86.70\%$&  $96.20\%$&  $86.40\%$ &  $91.10\%$ &	$98.30\%$ &	$90.56\%$\\ \hline 
         \myconvit&  B &  $85.20\%$ &  $95.50\%$ &  $84.87\%$ &  $89.90\%$ &  $97.90\%$ & $89.47\%$ \\ \hline 
         \myvitae&  B\_v2 &  $\textbf{89.80\%}$ &  $\textbf{97.50\%}$ &  $\textbf{89.53\%}$ &  $\textbf{93.60\%}$ &  $\textbf{98.80\%}$ & $93.29\%$ \\ \hline
    \end{tabular}
    \caption{AvgACC top1/top5, and F1score on the validation-split of Coleoptera\_Obs and Odonata\_Obs. }
    \label{tab:qualit_acc_f1}
\end{table*}
\begin{table*}
    \centering
    \begin{tabular}{|c|c|c|c|c|c|c|} \hline 
 
& \multicolumn{3}{|c|}{Coleoptera\_Art}& \multicolumn{3}{|c|}{Odonata\_Art}\\ \hline 
         Model&  Top 1&  Top 5&  F1 &  Top 1&  Top 5& F1 \\ \hline 
         \myincpt&  $65.20\%$&  $81.90\%$&  $65.56\%$&  $86.30\%$&  $95.10\%$& $86.11\%$\\ \hline 
         \myeffn&  $67.30\%$&  $82.90\%$&  $67.68\%$&  69.20\%&  $75.20\%$& $68.51\%$\\ \hline 
         \myresNet&  $65.30\%$&  $82.30\%$&  $65.81\%$&  $87.50\%$&  $95.40\%$& $87.50\%$\\ \hline 
         \myttv&  $66.80\%$&  $83.40\%$&  $67.16\%$&  $87.40\%$&  $96.30\%$& $87.49\%$\\ \hline 
         \mydeit&  $62.70\%$&  $79.60\%$&  $62.80\%$&  $85.90\%$&  $95.00\%$& $86.06\%$\\ \hline 
         \myvitdvitae&  $64.20\%$& $ 79.60\%$&  $62.80\%$&  $86.90\%$&  $96.30\%$& $86.67\%$\\ \hline 
         \myvitdeffn&  $66.20\%$&  $82.50\%$&  $66.66\%$&  $86.70\%$&  $96.20\%$& $86.33\%$\\ \hline 
         \myconvit&  $63.30\%$&  $80.30\%$&  $63.63\%$&  $85.40\%$&  $95.50\%$& $85.22\%$ \\ \hline 
         \myvitae& $\textbf{ 69.30\%}$& $ \textbf{84.80\%}$&  $\textbf{69.73}\%$&  $\textbf{90.60\%}$&  $\textbf{96.80\%}$& $\textbf{90.55\%}$ \\ \hline
    \end{tabular}
    \caption{AvgACC top1/top5, and F1score on datasets from Artportalen.}
    \label{tab:class_perf_art}
\end{table*}
\section{Results}
\subsection{Classification Performance}\label{Sec:class_perm}
 We first consider the average per-species of the metrics, then the distribution of the performance versus the amount of data available per-species, and we discuss the different behaviour of the models relative to the long tail distribution of the species. In the latter case, we discuss how the rarity of the species (the lack of samples) affects the performance of each model. Finally, we test the robustness of the models on the Artportalen.se dataset to observe the robustness of the models in generalising the new conditions.

\subsubsection{Results per-dataset}\label{ssec:results_dataset}
Tab.~\ref{tab:qualit_acc_f1} shows the results obtained in the evaluation with the test datasets of Observation.org. The Coleoptera\_Obs dataset is heavily unbalanced with a high number of species and a low number of samples. There is a minimal difference in performance on top-1, top-5 and F1score among \myvitae, \myeffn, and \myttv. It is interesting to observe that the \myttv model reaches a similar performance of \myvitae and \myeffn while having a less complex structure. All the other models show an accuracy lower by $2\%$. In particular, the distilled models \myvitdvitae and \myvitdeffn reach a plateau in the training phase after 30-40 epochs and they are not able to improve. That is also visible in the results shown in Tab.~\ref{tab:qualit_acc_f1}. On the contrary, \mydeit reaches high accuracy. The Odonata\_Obs dataset is also unbalanced but they consist of a high number of samples and a low number of species. We observe a similar behaviour with this dataset, where \myvitae, \myeffn, and \myttv obtained similar performances outperforming the other models. In particular, the \myvitae and the \myeffn outperform the others in all metrics for both datasets. We can conclude that these models can be better candidates compared to the other models on the average species accuracy if the focus is the accuracy performance. 

\begin{table}
    \centering
    \begin{tabular}{|c|c|c|c|} \hline 
 & \multicolumn{3}{|c|}{Coleoptera\_Obs}\\ \hline 
         model& Top 1&  Top 5&  F1 \\ \hline 
         ViT&  $81.70\%$&  $93.70\%$&  $81.42\%$\\ \hline 
         \myvitdvitae&  $ 86.60\%$&  $96.10\%$&  $86.11\%$ \\ \hline 
         \myvitdeffn&  $\textbf{86.70\%}$&  $\textbf{96.20\%}$&  $\textbf{86.40\%}$ \\ \hline
         \mydeit&  $84.10\%$ &  $95.10\%$ &  $83.90\%$\\ \hline  
& \multicolumn{3}{|c|}{Odonata\_Obs}\\ \hline
model& Top 1&  Top 5&  F1 \\ \hline 
        ViT& $88.80\%$  &$97.5\%$  & $88.40\%$\\ \hline 
        \myvitdvitae& $90.90\%$&  $98.30\%$& $90.33\%$\\ \hline
        \myvitdeffn& $\textbf{91.10\%}$ &	$\textbf{98.30\%}$ &	$\textbf{90.56\%}$ \\ \hline
        \mydeit&  $89.70\%$ &  $97.70\%$ & $89.32\%$ \\ \hline
    \end{tabular}
    \caption{AvgACC top-1/top-5, and F1 score on Coleoptera\_Obs and \\Odonata\_Obs for the ablation study on ViT models.}
    \label{tab:ViT_Focus}
\end{table}

\paragraph{Ablation study on knowledge distillation} 
We wished to understand the contribution of knowledge distillation to the model and so compared these against the same Vit trained in knowledge distillation. Tab.~\ref{tab:ViT_Focus} shows the results of vanilla ViT compared to the other models obtained by knowledge distillation. We observe that the models trained in knowledge distillation outperform the vanilla ViT. That can be explained by the fact that the datasets are small and can not alone satisfy the needs of ViT, but together with the distillation of knowledge from models with high performance we can obtain better results. For this reason, in this paper, we consider the \myvitdeffn and \myvitdvitae rather than the vanilla ViT.

\paragraph{Results on Artportalen dataset}
In Sec.~\ref{ssec:results_dataset}, we evaluated the models on the test set of the Observation.org dataset. The test set belongs to the same distribution as the train split even if the images are completely unknown to the models. To evaluate their robustness, we evaluate the models computing the avgACC top-1 and top-5 and F1score with Coleoptera\_Art and Odonata\_Art (Tab.~\ref{tab:class_perf_art}). With Coleoptera\_Art, the overall performance drops by 20\% compared to the one observed for Coleoptera\_Obs. This behaviour underlines a lack of robustness of all these models to different distributions. Among the models, we observe that all the models reach 64\% on average for top-1 and 80\% for top-5, while F1score is on average lower than 65\%. The performance follows the pattern observed with Coleoptera\_Obs in terms of the single performance of each model. Among the CNN group, the \myeffn reaches the better performance on all the metrics, in the ViT group the \myttv shows higher results and in the LBVT group, the \myvitae outperforms the others. There is a drop in performance also with Odonata\_Art, even if it is limited to 10\% and only with \myeffn. All the other models show robust behaviour with the distribution of Artportalen, with performance that is in line with the one obtained with Odonata\_Obs. Among the CNN group, the \myresNet shows high performance, in the ViT group, the \myttv reaches the better one, and in the LBVT the \myvitae is the model with the highest performance. The \myvitae is the model which outperforms all the other models in both datasets and is a good candidate if we consider classification performance.

\begin{figure*}
\centering
  \includegraphics[width=\linewidth]{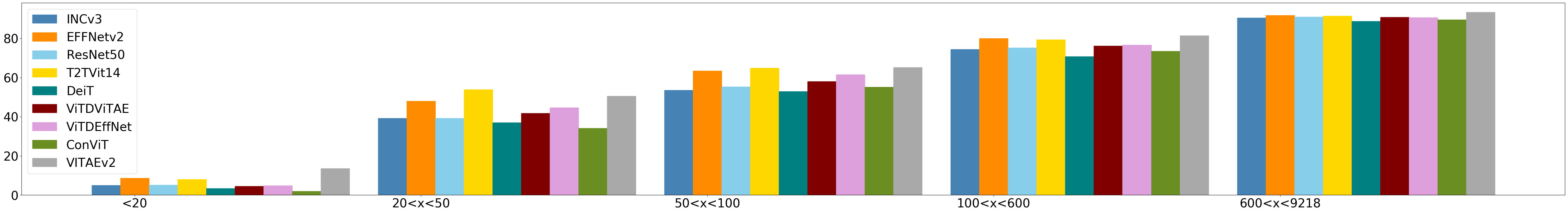}
  \caption{Coleoptera\_Obs}
  \label{fig:bar_species_Coleo}
\end{figure*}
\begin{figure*}
  \centering
  \includegraphics[width=\linewidth]{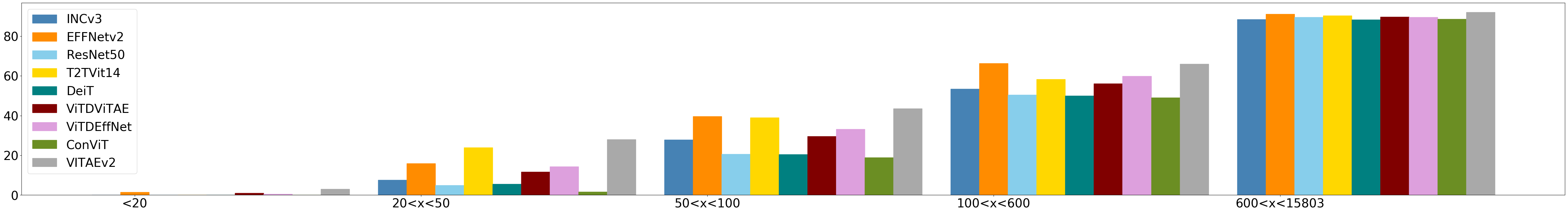}
  \caption{Odonata\_Obs}
  \label{fig:bar_species_Odo}
\caption{Top-1 results obtained per species with each model for the species-wise groups (x-axis) based on the number of samples available in the training split. From left to right, from the most rare species to the most common one.}
\label{fig:bar_species}
\end{figure*}
\begin{figure*}
\centering
\begin{subfigure}{.5\textwidth}
  \centering
  \includegraphics[width=.9\linewidth]{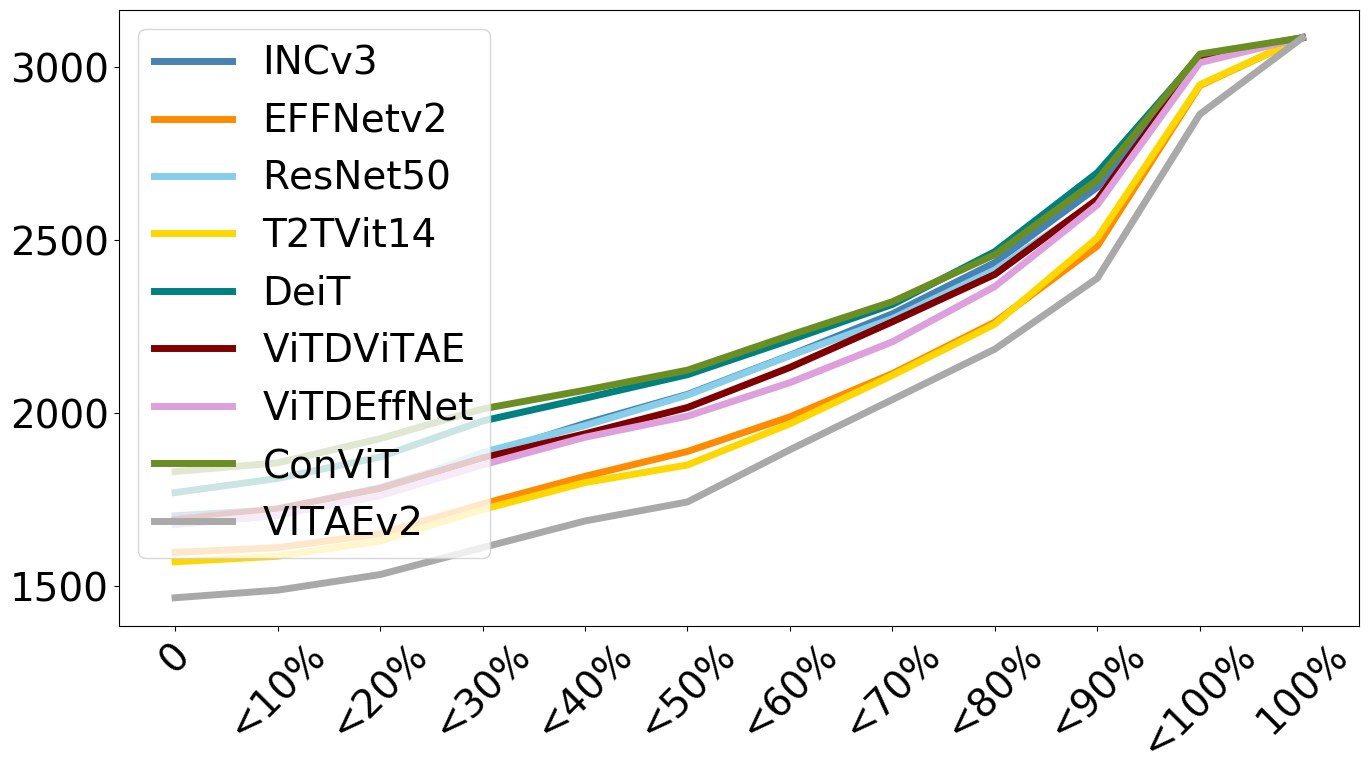}
  \caption{Coleoptera\_Obs}
  \label{fig:line_species_Coleo}
\end{subfigure}%
\begin{subfigure}{.5\textwidth}
  \centering
  \includegraphics[width=.9\linewidth]{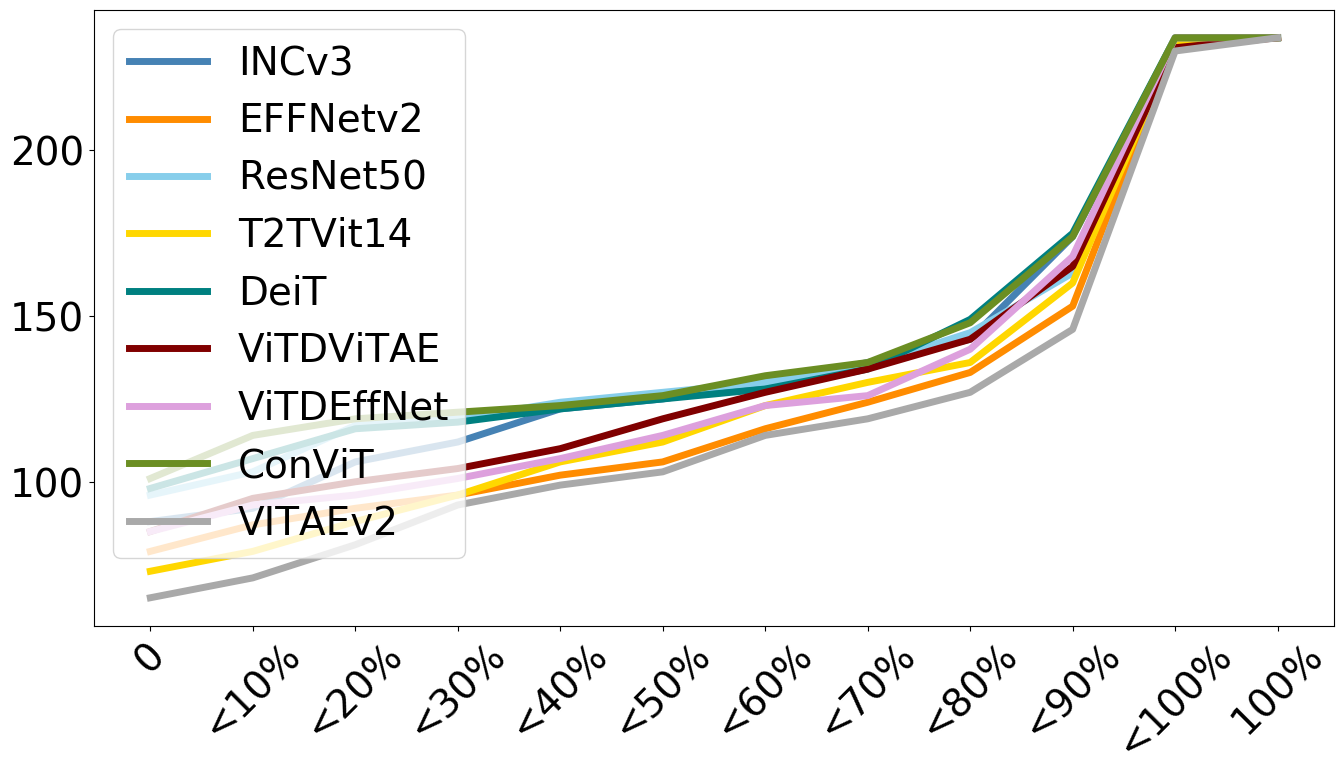}
  \caption{Odonata\_Obs}
  \label{fig:line_species_Odo}
\end{subfigure}
\caption{The lines show the number of species (x-axis) by the accuracy top-1 (x-axis) obtained by each model. E.g.: in Fig.~\ref{fig:line_species_Coleo} the \myeffn classifies almost 2100 species with avgACC top-1 between $60\%$ and $70\%$.}
\label{fig:line_species}
\end{figure*}
\begin{figure*}
\centering
\begin{subfigure}{.5\textwidth}
  \centering
  \includegraphics[width=.9\linewidth]{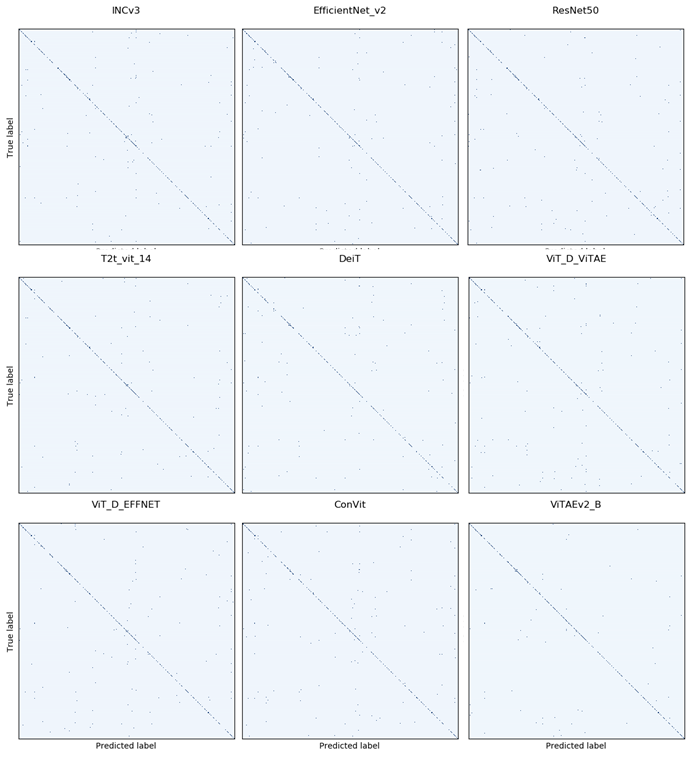}
  \caption{Coleoptera\_Obs}
  \label{fig:conf_matrix_Coleo}
\end{subfigure}%
\begin{subfigure}{.5\textwidth}
  \centering
  \includegraphics[width=.9\linewidth]{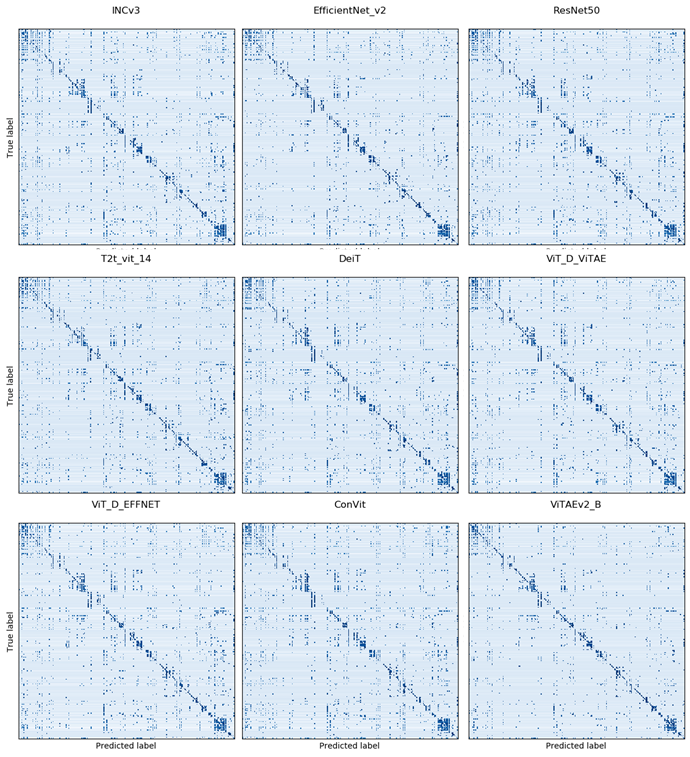}
  \caption{Odonata\_Obs}
  \label{fig:conf_matrix_Odo}
\end{subfigure}
\caption{Confusion matrices per order and models. We do not report the names of species in figures to maintain the matrices readable. The classes are ordered by taxonomic name, which, in this paper, is a pair <genus\_species>.}
\label{fig:conf_matrix}
\end{figure*}

\begin{figure*}
\begin{subfigure}{.5\textwidth}
  \centering
  \includegraphics[width=\linewidth]{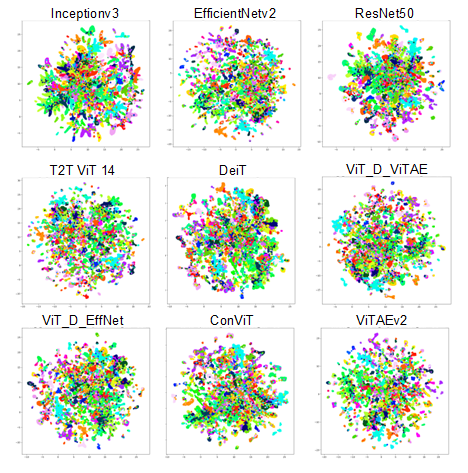}
  \caption{Coleoptera\_Obs}
  \label{fig:UMAP_Coleo}
\end{subfigure}
\begin{subfigure}{.5\textwidth}
  \centering
  \includegraphics[width=\linewidth]{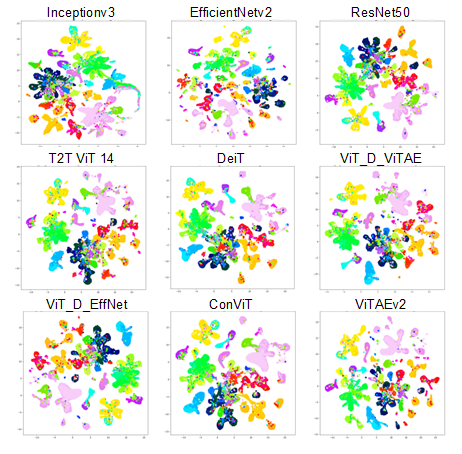}
  \caption{Odonata\_Obs}
  \label{fig:UMAP_Odo}
\end{subfigure}%
\caption{UMAP visualisation of the embedding space generated at the species level by each of the trained models.}
\label{fig:UMAP}
\end{figure*}

\subsubsection{Results per-species}
The results presented in Sec.~\ref{ssec:results_dataset} offer insight into the effect of introducing transformers, which outperform CNN models in accuracy. The fine-grained classification tasks are focused on what is happening at the class level. Therefore, we examine the accuracy distribution among species and how the models perform when classifying rare species. In Fig.~\ref{fig:bar_species}, the bar charts show the top-1 accuracy divided according to the range of available data in the training split for a single species on the x-axis. The models have been trained on the original distribution of the dataset without balancing the species to test the ability of the model to recognise rare species.  We consider the species with less than 20 samples, as rare species and their rarity is identified with the lack of data in the dataset. We notice a difference in the model's behaviour between the Coleoptera\_Obs and Odonata\_Obs case studies. In Fig.~\ref{fig:bar_species_Coleo}, all models reach a mean accuracy lower than 20\% with rare species. Among them, the \myvitae has an accuracy between 15\% and 20\% while all the other models' accuracy is lower than 15\%. With species with less than 100 samples in training, the performance of all the models grows, but only \myeffn, \myttv, and \myvitae reach almost 60\% while all the others are below 50\%. The \myeffn achieves competitive results that are slightly lower than the models based on transformers. In Fig.~\ref{fig:bar_species_Odo}, the behaviour of the models is not similar to the Coleoptera\_Obs case study. With rare species, only \myeffn and \myvitae show any ability to learn the class characteristics. The performance increases in all the models while increasing the number of samples in training for species. with lower than 600 samples per species, the \myvitae always shows better performance followed by \myeffn and \myttv. With the high amount of data, almost all the models can reach good results, only \myvitdeffn and \myvitdvitae have poor results.
In Fig.~\ref{fig:line_species}, we display the number of species each model classifies within a given range of avgACC top-1. For the Coleoptera\_Obs, \myvitae has the lowest number of species with 0\% accuracy while the \myconvit has the highest number of species with 0\% accuracy. With \myvitae, \myeffn, and \myttv almost 2000 species have an accuracy lower than 70\%, while, with all the other models, there are around 2500 species. \myvitae has the highest number of species with 100\% accuracy, around 200 species while all the other models have less than 100 species with the same accuracy. With Odonata\_Obs, we observe a similar trend in the performances of the models. \myvitae, \myeffn, and \myttv have a low amount of species with 0\% accuracy, in particular for \myvitae with have less than 20 species. For all the models, around 160 species have an accuracy lower than 90\%. 
In Odonata\_Obs and Coleoptera\_Obs, we observed a different behaviour based on the number of data in training but a similar behaviour on the number of the species recognised. The \myvitae proves to be less sensitive to the low amount of data available ending with fewer species with $0\%$ top-1. Even if the other models are more sensitive to the data available in the training dataset, they reach good performance.
Finally, we observe the confusion matrices, to evaluate the performance of the models based on the number of true positives, true negatives, false positives, and false negatives. Fig.~\ref{fig:conf_matrix} shows the confusion matrices with the test datasets of Coleoptera\_Obs and Odonata\_Obs. In the confusion matrix, the rows/columns close to each other are species which share the genus. For Coleoptera\_Obs, we do not observe visible clusters. The predictions provided by the models highlight the diagonal of the matrices, which is supported by the high overall accuracy that all the models obtained. By a closer look, it is visible that the \mydeit, \myresNet, \myincpt, and the \myconvit show many predictions not lying on the diagonal. This is supported by the F1score metrics shown in Tab.~\ref{tab:qualit_acc_f1}, where these three models present lower results. With Odonata\_Obs, there are some visible clusters shared among models, which denote confusion among species from the same genus i.e. the difficulty of the fine-grained task. This underlines how all these models behave similarly for some classes. Also in this case study, the \myconvit, \mydeit, \myresNet, and \myincpt show more prediction far from the diagonal.
\begin{table*}
    \centering
    \begin{tabular}{|c|c|l|l|c|c|} \hline 
 \multicolumn{2}{|c|}{}&   \multicolumn{4}{|c|}{Silhouette score}\\ \hline 
 &embedding&   \multicolumn{2}{|c|}{Coleoptera\_Obs}&\multicolumn{2}{|c|}{Odonata\_Obs}\\ \hline 
         Model &  dim&    genus&species 
&genus&  species \\ \hline 
         \myincpt &  [1,1024]  &    0.1293&0.1239  
&0.0407&  0.0130   \\ \hline 
         \myeffn &  [1,1024]  &    0.1333&0.1577 
&0.1669&  0.2165  \\ \hline 
         \myresNet &  [1,2048] &    0.0439&0.0491 
&0.1453&  0.1448  \\ \hline 
         \myttv &  [1,384]  &    0.1502&0.1739 
&0.1716&  0.2345 \\ \hline 
         \mydeit &[1,768]&   0.1217&0.1039&0.1039
& 0.1704 \\ \hline
         \myvitdvitae &  [1,768]  &   0.1403 &0.1469 
&0.1844 & 0.2258 \\ \hline          
         \myvitdeffn&  [1,768]  &    0.1518&0.1689
&\textbf{0.2063}&  0.2482\\ \hline 
         \myconvit &  [1,768]  &    0.1337&0.1147 
&0.1855&  0.2157\\ \hline 
         \myvitae & [1,1024]&   \textbf{0.2189}&\textbf{0.2517} &0.1925& \textbf{0.3217}\\ \hline
    \end{tabular}
    \caption{Embedding dimension for each model and average Silhouette score\\ obtained on the Odonata\_Obs and Coleoptera\_Obs embedding space.}
    \label{tab:Sil_score}
\end{table*}
\subsection{Embedding performance}\label{Sec:embed_perm}
We provide a quantitative and visual analysis of the embedding spaces learned by the trained models. Fig.~\ref{fig:UMAP} shows the visualisation of the embedding space of each model for the Observation.org datasets, obtained with UMAP. In Fig.~\ref{fig:UMAP_Coleo}, the embedding space aims to distinguish among 3087 classes while in Fig.~\ref{fig:UMAP_Odo}, the classes are 235. We guide the evaluation of the visual representation taking into consideration the quantitative evaluation of the embedding space with the silhouette score. In Tab.~\ref{tab:Sil_score}, we separately report the mean silhouette score considering the genus and species levels. For the Coloeptera\_Obs, the \myresNet obtains the lowest (worst) results both on genus and species and, in \ref{fig:UMAP_Coleo}, the embedding space of \myresNet is collapsed in a central focal point. The \mydeit, \myconvit and \myincpt obtain a higher score, and also in these cases the embedding spaces appear collapsed on lateral focal points. The \myeffn, \myttv, \myvitdeffn, and \myvitdvitae obtained similar results, while \myvitae outperforms all the other models at genus and species levels. These results are confirmed by the visual analysis where the embedding spaces appear distributed in the space. With Odonata\_Obs in Fig.~\ref{fig:UMAP_Odo}, the \myincpt obtained the lowest results at both levels and the embedding space shows multiple overlaps among species and does not present a structure. \myconvit, \myeffn, and \myttv obtained good results at the species level, the overlapping is less present and the points look spread in the space. Finally, \myvitae outperforms all the models at the species level and obtained good results together with \myvitdeffn at the genus level where there is a minimum overlap, but the overall results in a good distancing of the species. Combined with the visual analysis, the quantitative analysis suggests that the \myttv and \myvitae generate embedding spaces with a better distribution of embedding in the space. 

\begin{table*}
    \centering
    \begin{tabular}{|c|c|c|c|c|c|c|c|c|c|} \hline 
          &\multicolumn{3}{|c|}{NN Structural Information}&     \multicolumn{3}{|c|}{Coleoptera}&\multicolumn{3}{|c|}{Odonata}\\ \hline 
         \multirow{2}{*}{Model}&  \#layers &FLOPS&  Venue& Inf-time&Param &Train-time&Inf-time&  Param &Train-time \\ 
         && (G)&&(ms)&(M)& (ms)& (ms)&(M)& (ms) \\ \hline 
         \myincpt &  48  &5.71&  CVPR'16&     14.6&28.1&97.2 &18.0&  22.3&87.1\\ \hline 
         \myeffn &     154& 24.0&  ICML’21&     28.2&57.7&181.1 &27.9&  51.6&168.2\\ \hline 
         \myresNet &     50 & \textbf{3.80}&  CVPR’16&     10.2&29.8&52.5 &\textbf{8.3}&  24.0&44.8\\ \hline
         \myttv &     16 & 5.20&  ICCV’21&     10.5&\textbf{22.3}&52.6 &8.4&  \textbf{21.2}&45.7\\\hline
        \mydeit& \textbf{12}& 17& ICML’21&    \textbf{6.0}&90.5&\textbf{31.3} &34.9& 86.2& \textbf{35.3}\\ \hline 
         \myvitdvitae&  \textbf{12}  &4.60&  ICML’21&     25.7&22.9&53.0 &25.5&  21.8&82.1\\ \hline 
         \myvitdeffn&  \textbf{12}& 4.60&  ICML’21&     28.9&22.9&60.0 &24.2&  21.8&87.8\\ \hline 
         \myconvit &   32 & 17.0&  ICML’21&     17.2&88.1&67.6 &11.2&  86.0&55.8\\ \hline 
         \myvitae &   41 &24.3&  IJCV'22&     24.3&91.8&165.6 &24.4&  88.9&151.5\\ \hline 
         \multicolumn{4}{|c|}{}&    \multicolumn{6}{|c|}{input - [1,3,224,224]}\\ \hline
    \end{tabular}
    \caption{Structural information, inference/training time, and the number of parameters of each model for Odonata\_Obs and Coleoptera\_Obs case studies. }
    \label{tab:comp_cost}
\end{table*}

\begin{figure*}
\begin{subfigure}{.5\textwidth}
  \centering
  \includegraphics[width=1.05\linewidth]{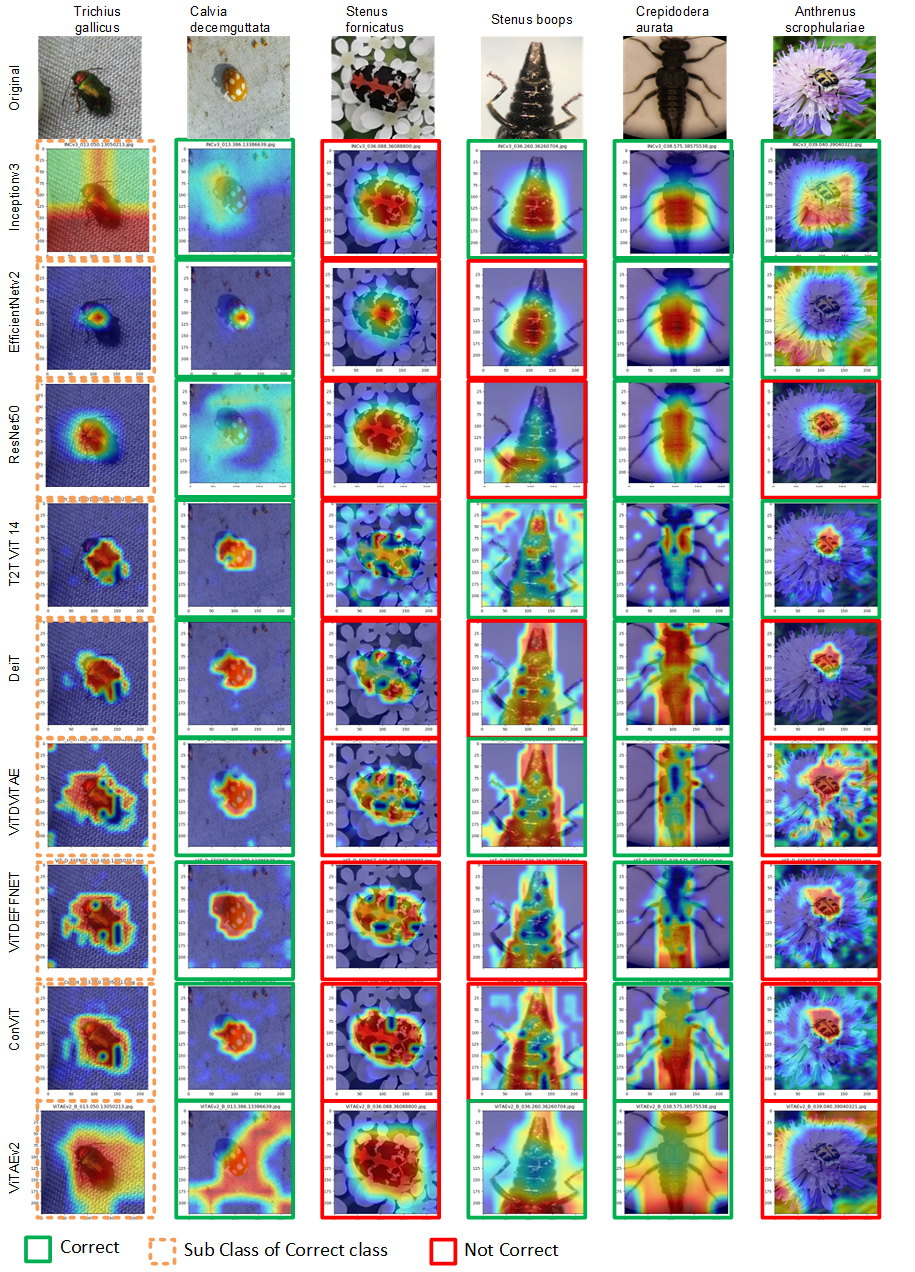}
  \caption{Coleoptera\_Obs}
  \label{fig:gradCam_Coleo}
\end{subfigure}
\begin{subfigure}{.5\textwidth}
  \centering
  \includegraphics[width=.9\linewidth]{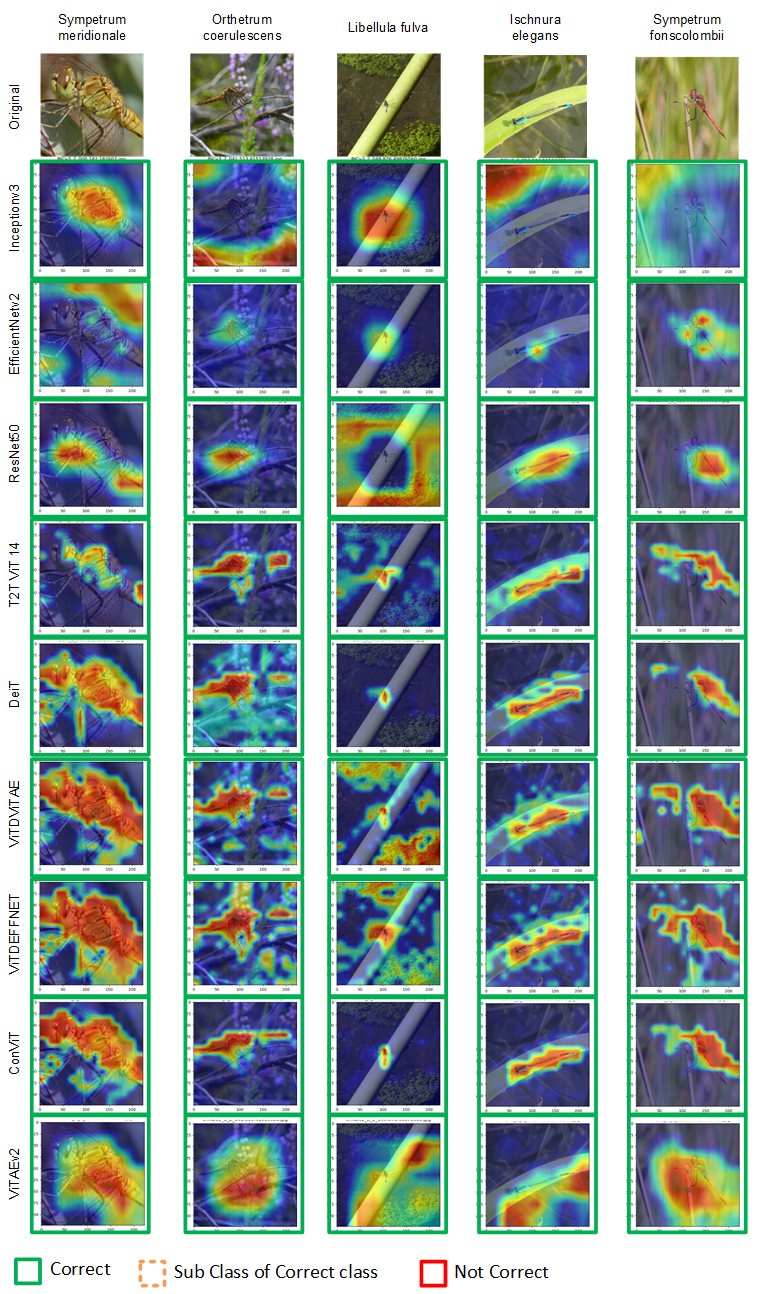}
  \caption{Odonata\_Obs}
  \label{fig:gradCam_Odo}
\end{subfigure}%
\caption{Visualisation of gradient activation with GradCam. In the first row, there are the original inputs presented to the models; in each row, the heatmaps of the gradient activity show the regions with the higher (red) and lower (blue) influence on the decision made by the model.}
\label{fig:grad_cam_visualisation}
\end{figure*}

\subsection{Computational cost performance}\label{Sec:cost_perm}
We compute the metrics for each model without any routine such as data augmentation or optimisation strategy. In Tab.~\ref{tab:comp_cost}, we observe that the number of layers strongly correlates with the model's group. The CNN have a deeper structure compared to equally complex fully/partially-transformers models. This is not surprising considering the characteristics of convolutional models versus transformers. Convolutional models require more layers to extract global features, while transformers can extract global features with just a few layers using patches. The \myresNet outperforms all the others in terms of FLOPS, followed by the \myvitdeffn and \myvitdvitae which refer to the ViT model used as a student, and \myttv. These four models are the smaller and more portable models analysed in this paper. The two case studies are presented separately. Each has a different head dimension, corresponding to a different number of parameters. The \myvitae requires almost 2$\times$ the number of trainable parameters for \myeffn, while if we consider the training time, the \myeffn takes slightly more time than \myvitae. The \myttv and \myresNet are the fastest models among the ones considered. Finally, the inference time manifests the same behaviour as described for the training time.

\subsection{Gradients activity performance}\label{Sec:gradCam_perm}
The visualisation of the gradient activity computed with GradCam is presented in Fig.\ref{fig:grad_cam_visualisation} with samples from Coleoptera\_Obs and Odonata\_Obs. We select inputs from different species, with different resolutions to observe how the models react to these variations. In each row, we present a heatmap of the gradient activity at the final layers of each model, as described in Sec.~\ref{Sec:gradCam_metrics}, and we map the heatmap on the input images to identify the areas of interest for the model. The heatmaps show the regions with the higher and lower influence on the decision made by the model respectively in red and blue. In Fig.~\ref{fig:gradCam_Coleo}, the samples are of six distinctive species of Coleoptera, the third and the sixth present a botanical background while all the others present a concrete background. Overall, the models have different gradient focus, we can observe similarities among models from the same group. Models in the CNN group show areas of interest which are relative to the feature maps which are highly responsive during the classification. Models in the ViT group are more focused on the detailed silhouette of the subject. The two models considered for LBVT have different behaviour, while \myconvit has a focus similar to models from the ViT group, the \myvitae focus is not limited to the subject but also part of the background is taken into consideration. From the coloured and dashed squares, we observe that some of the samples are misclassified by all the models, in particular with the first, and third images. All the models identify the first image with the class \textit{Trichius gallicus gallicus}, while the class labelled in the data is \textit{Trichius gallicus}. This behaviour can be due to two main factors: 1- the two species are closely related and present a similar morphology; 2- in the training dataset the \textit{Trichius gallicus} consists of 427 samples while the \textit{Trichius gallicus gallicus} has 3823 samples. Hence, the models tend to choose the more common species. The third image is misclassified by all the models, we do not observe a pattern shared among all of them. \myeffn, \myincpt, \myttv, and \myvitdvitae misclassified the image for a species from the same genus \textit{Stenus ater}. All the other models misclassify the image with classes unrelated to genus or subspecies level. Finally the last image with species \textit{Anthrenus scrophulariae} appears to be confusing for the majority of the models which focus on the flower instead of the insect.

Fig.~\ref{fig:gradCam_Odo} shows five samples of Odonata\_Obs, the images show different species and the animals are at different distances from the camera. All the samples presented are well classified by all the models considered. The figure shows how the models focus differently while predicting the correct class. We observe that the overall behaviour of the models is similar to that observed with Coleoptera. With models in CNN, the focus areas are not always on the subject but rather on the background. The models in the ViT group and the \myconvit model show a well-defined focus on the subject of the images. Finally, \myvitae focuses on the subject and background to compute the classification.

\begin{table*}
    \centering
    \begin{tabular}{|c|c|c|c|c|c|c|c|c|}\hline
 \multicolumn{9}{|c|}{Coleoptera\_Obs}\\\hline \hline 
        
         Model&  C: Top-1&  SC: genus&SC: species&  Inf T& Train T&mem space&  GA details&  GA area\\ \hline 
         \myincpt & \neutral & \neutral & \neutral &  \neutral   & \firstminus& \neutral &  \secondminus&  \thirdplus\\ \hline 
         \myeffn &  \thirdplus&   \neutral  & \neutral &   \secondminus& \thirdminus& \neutral &  \firstminus&  \secondplus\\ \hline 
         \myresNet &  \neutral   &   \thirdminus&\thirdminus&   \secondplus& \secondplus& \neutral &  \thirdminus&  \neutral  \\ \hline 
         \myttv &  \secondplus&   \thirdplus&\secondplus&   \thirdplus&  \neutral  &\firstplus&  \neutral   &   \neutral  \\ \hline 
         \mydeit&  \thirdminus&   \secondminus&\secondminus&   \firstplus& \firstplus&\secondminus&   \neutral  &   \neutral  \\ \hline 
         \myvitdvitae&  \firstminus&   \neutral   &  \neutral &   \firstminus& \thirdplus&\secondplus&  \secondplus&  \secondminus\\ \hline 
         \myvitdeffn&  \neutral   &   \secondplus&\thirdplus&   \thirdminus&  \neutral  &\secondplus&  \firstplus&  \firstminus\\ \hline
         \myconvit & \secondminus&  \firstminus&\firstminus&  \neutral   &  \neutral  &\firstminus& \thirdplus&\thirdminus\\\hline
         \myvitae & \firstplus& \firstplus&\firstplus&  \neutral   & \secondminus&\thirdminus&  \neutral  &\firstplus\\\hline\hline
         \multicolumn{9}{|c|}{Odonata\_Obs}\\\hline \hline 
         \myincpt &  \secondminus&   \thirdminus&\thirdminus&   \neutral   &  \neutral  & \neutral  &  \secondminus&  \thirdplus\\ \hline 
         \myeffn &  \secondplus&   \neutral   &  \neutral &   \thirdminus& \thirdminus& \neutral  &  \firstminus&  \secondplus\\ \hline 
         \myresNet &  \neutral   &   \secondminus&\secondminus&   \firstplus& \secondplus&  \neutral &  \thirdminus&  \neutral   \\ \hline 
         \myttv &  \thirdplus&   \thirdplus&\thirdplus&   \secondplus& \thirdplus&\firstplus&  \neutral   &  \neutral   \\ \hline 
         \mydeit&  \thirdminus&   \firstminus&\firstminus&   \neutral   & \firstplus&\secondminus&   \neutral  &  \neutral   \\ \hline 
         \myvitdvitae&  \neutral   &   \neutral   & \neutral  &   \secondminus&  &\secondplus&  \secondplus&  \secondminus\\ \hline 
         \myvitdeffn&  \neutral   &   \firstplus&\secondplus&   \neutral   & \firstminus&\secondplus&  \firstplus&  \firstminus\\ \hline
         \myconvit & \secondminus&  \secondplus& \neutral  &  \thirdplus&  \neutral  &\firstminus& \thirdplus&\thirdminus\\\hline
         \myvitae & \firstplus& \secondplus&\firstplus&  \firstminus& \secondminus&\thirdminus&  \neutral  &\firstplus\\\hline
    \end{tabular}
    \caption{Summery of the metrics discussed in the paper for an overall visualisation: avgACC top-1, Silhouette score (SC) at genus and species levels, inference and train time (inf/train T), number of parameters (mem space), gradient activation (GA) on details, and areas. Symbols: \{$+++$:best; $++$:second best; $+$:third best; $=$:average; $-$: third to worst; $--$: second to worst; $---$: worst\}.}
    \label{tab:wrap_up}
\end{table*}
\begin{figure*}
\begin{subfigure}{.5\textwidth}
  \centering
  \includegraphics[width=\linewidth]{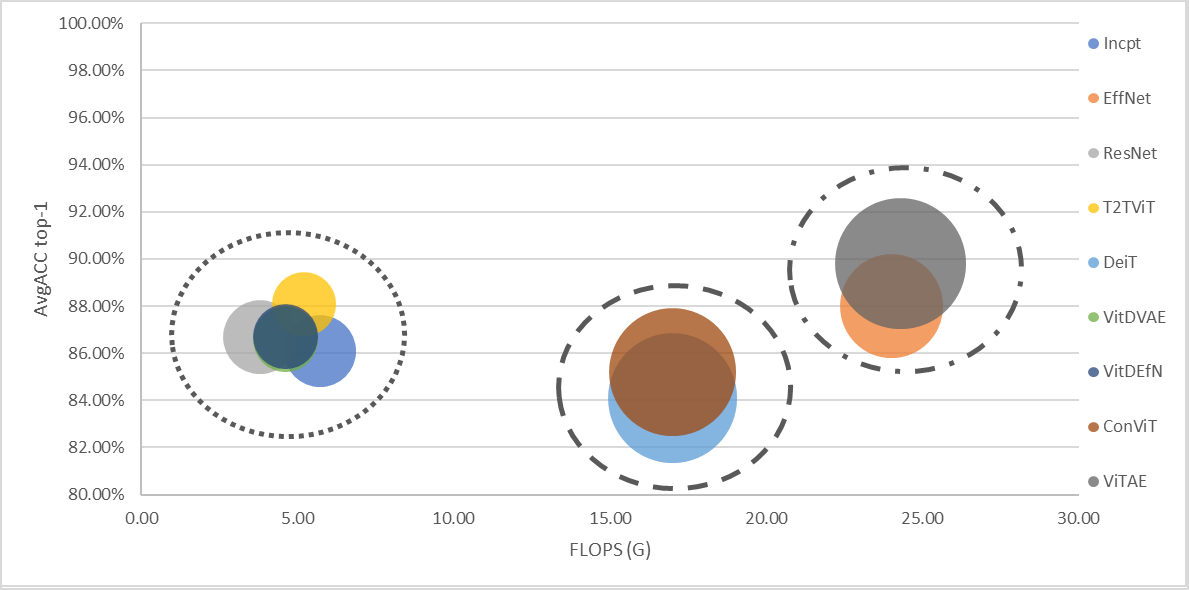}
  \caption{Coleoptera\_Obs}
  \label{fig:ACCvsFLOPS_Coleo}
\end{subfigure}
\begin{subfigure}{.5\textwidth}
  \centering
  \includegraphics[width=\linewidth]{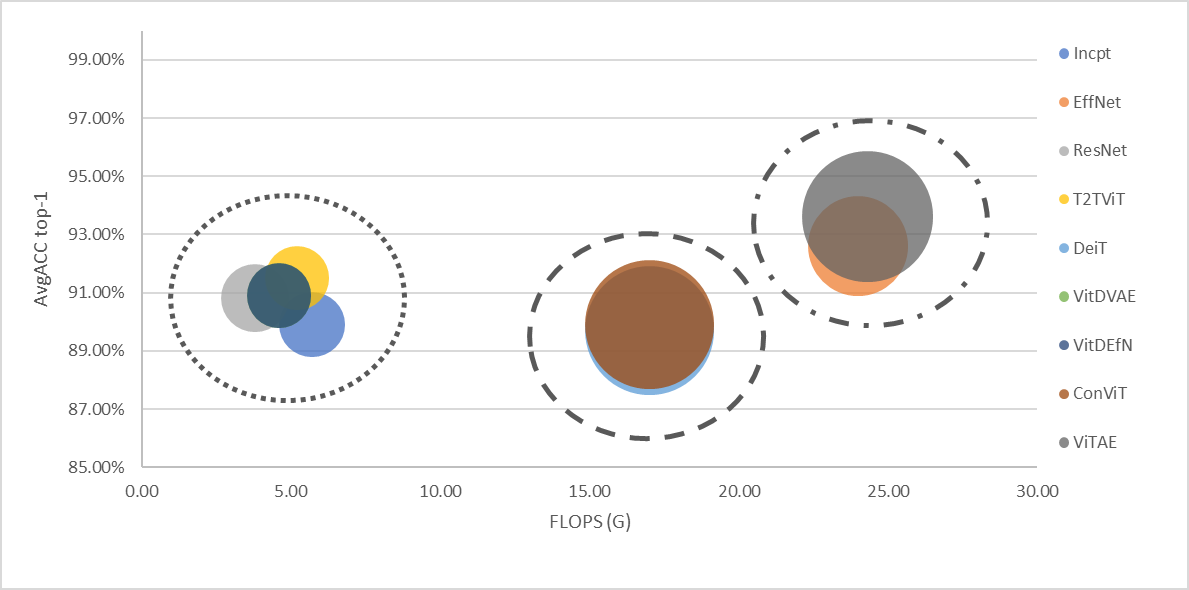}
  \caption{Odonata\_Obs}
  \label{fig:ACCvsFLOPS_Odo}
\end{subfigure}%
\caption{Contextualise FLOPS demand with avgACC top-1 of each model. The dimension of the circles expresses the number of parameters (M) required by the models, Tab.~\ref{tab:comp_cost}. The x-axis is the FLOPS(G) shown in the Tab.~\ref{tab:comp_cost} while the y-axis is the avgACC top-1 shown in the Tab.~\ref{tab:qualit_acc_f1}.}
\label{fig:ACCvsFLOPS}
\end{figure*}
\begin{figure*}[!t]
\begin{subfigure}{.5\textwidth}
  \centering
  \includegraphics[width=\linewidth]{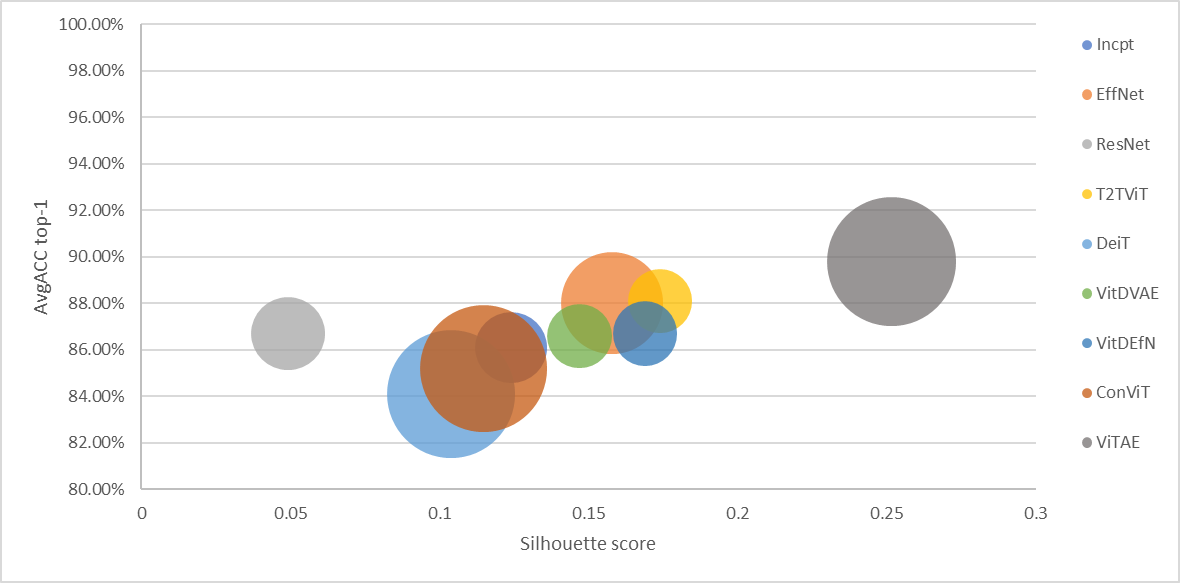}
  \caption{Coleoptera\_Obs}
  \label{fig:ACCvsSC_Coleo}
\end{subfigure}
\begin{subfigure}{.5\textwidth}
  \centering
  \includegraphics[width=\linewidth]{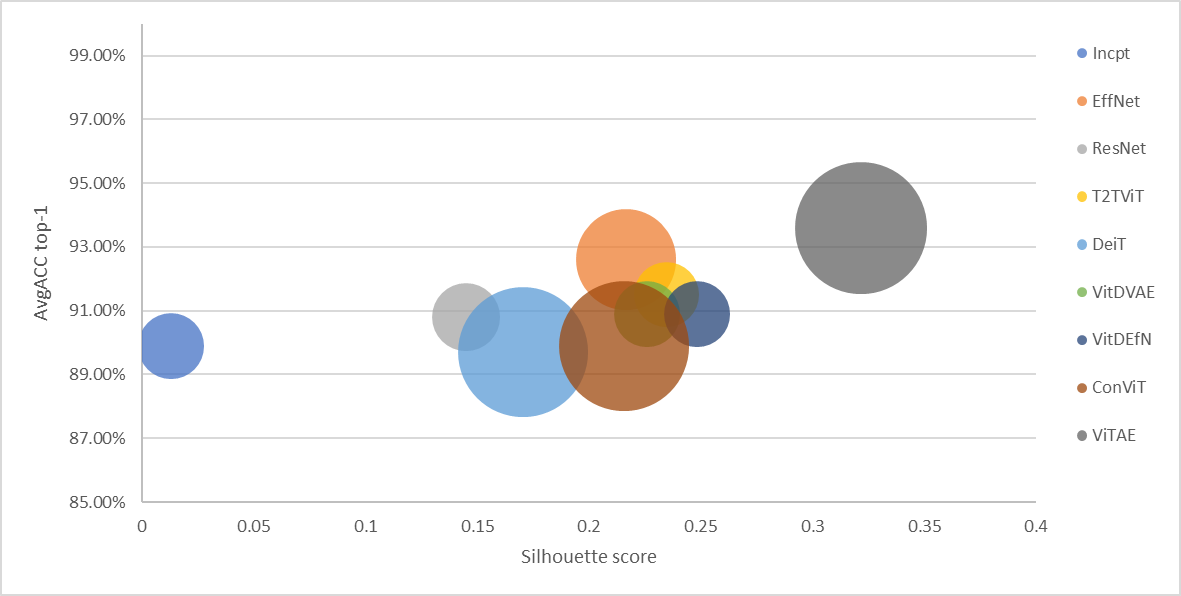}
  \caption{Odonata\_Obs}
  \label{fig:ACCvsSC_Odo}
\end{subfigure}%
\caption{Contextualise embedding space quality through Silhouette score results with avgACC top-1 of each model. The dimension of the bubbles is based on the number of parameters (as in Fig.~\ref{fig:ACCvsFLOPS}).}
\label{fig:ACCvsSC}
\end{figure*}

\section{Discussion}\label{sec:discussion}
In this section, we consider all the analyses presented in the paper and we present the overall view of the performance obtained by the models in the presented case study of fine-grained classification of images of insects. Tab.~\ref{tab:wrap_up} summarises the various metrics discussed in the paper. We use a dictionary of symbols to indicate the best and the worst models in each category. We distinguish between focus on areas and details, not as positive or negative aspects but more as the ability of the models. The type of focus is an aspect to evaluate in relation to the dataset of interest to answer the question: are the details that can distinguish the species or is it more the overall image that can help in the classification? From the Tab.~\ref{tab:wrap_up} we observe several characteristics of the models:
\begin{itemize}
    \item \myincpt doesn't show any particularly positive remarks in any of the two datasets. This model focuses on areas of the image.
    \item \myeffn is the third best model for avgACC top-1 and its gradient activation is on areas rather than details with both datasets. 
    \item \myresNet shows to be fast in inference and train time but with low results in embedding space generation. It has not a particular focus.
    \item \myttv has positive remarks on classification performance, embedding space generation, and inference and train time while using the lowest amount of memory. 
    \item \mydeit is the fastest among the models selected but with low performance in classification and embedding.
    \item \myvitdvitae has negative points on inference time and focus on areas of the pictures but it focuses on details and requires a low amount of memory space. This model has an evident focus on details while a low focus on wide areas.
    \item \myvitdeffn is one of the best models on the embedding space generation and the memory demand.
    \item \myconvit shows a high focus on details of the subject in the focus. On the Odonata\_Obs dataset, it has good performance on classification and inference time.
    \item  \myvitae has the best performance in classification and embedding space, with a wide focus on the subject and background. While its performance on computational cost is the lowest.
\end{itemize}
Among all the models, we identify \myvitae, \myeffn. and \myttv as the models with the best classification performance; \myvitdeffn, \myvitdeffn, and \myttv with the best embedding space; \myresNet and \myttv the fast models both at inference and training time; \myttv, \myvitdeffn, and \myvitdvitae as the smaller models.

Fig.~\ref{fig:ACCvsFLOPS} shows the FLOPS demand with the avgACC top-1 for each of the models considered in the paper. In both Odonata\_Obs and Coleoptera\_Obs, we identify three sets of models, highlighted in the Fig.~\ref{fig:ACCvsFLOPS_Odo}~\ref{fig:ACCvsFLOPS_Coleo} with circles. in the circle of dots and dash, there are models with the highest FLOPS demand and highest top-1, these models are the \myvitae and \myeffn. In the circle of the dash, some models do not show high accuracy and not a low FLOPS demand, \mydeit and \myconvit. Finally, in the circle of dots, there are all the other models which obtain good top-1, even if they do not outperform the others, and low FLOPS demand, these models are a good trade-off between cost and performance. In this last group, the \myttv has the highest top-1, and it is a good candidate in case we have limits on the computational cost.

Fig.~\ref{fig:ACCvsSC} present the avgACC top-1 and the Silhouette score (Sec.~\ref{Sec:embed_perm}) for each of the models. The behaviour of almost all the models is consistent among the two datasets. We observe that the behaviour of \myresNet in Fig.~\ref{fig:ACCvsSC_Coleo} and \mydeit in Fig.~\ref{fig:ACCvsSC_Odo} is not consistent, this can be due to the differences between the two datasets (number of species) or the limit of the network. There is no evidence of a connection between the number of parameters and the Silhouette score, in fact in both the figures the models do not show any related trend. We do notice a relationship between the avgACC top-1 and the Silhouette score: the growths of the avgACC top-1 and the Silhouette score are directly proportional. This observation suggests that we need models with higher embedding performance to obtain better results in the fine-grained classification. The Silhouette score directly evaluates the ability to form compact clusters in the embedding space. Hence, this gives credit to the Silhouette score as a metric to evaluate models applied for fine-grained classification. 

\section{Conclusion}\label{sec.:conclusion}
In the paper, we assess the performance of models at the fine-grained classification tasks for three groups of deep learning models: the convolutional neural network, the Vision Transformer, and the Locality-based vision transformer. For each of these groups, we consider models which at the state of the art are the newest with high classification performance or the most used ending up with nine models. For this study, we use the datasets of images of insects collected by citizen scientists and available on Observation.org~\cite{Obs}. At the state of the art, models from the convolutional neural network group are often applied without any investigation of models which belong to the other groups. We demonstrate that only consideration of the classification performance is not enough to evaluate a model in this delicate task, we need to consider its inner ability to deal with rare species, and how feasible it is to use that model in terms of computational requirement. Also, models trained on classification tasks are often used as the backbone for more specific tasks and with this perspective, the evaluation of the embedding space is an important aspect to consider.  We also show that models based on transformers can satisfy more of these aspects while obtaining higher accuracy and for that, they are good candidates for the fine-grained classification tasks. Finally, we need to consider the end use of these models on datasets of Odonata and Coleoptera and for these case studies we observe that if the performance and the robustness are the features mainly required, the \myvitae and \myeffn models are the most suitable for the fine-grained tasks with a preference for the \myvitae; if the focal point is for public use so the inference speed is to be considered, the \myttv demonstrated to achieve good performance faster than the others and it shows a promising trade-off between performance and costs. 
\section*{Acknowledgement}
This research was supported by the EU Horizon Europe projects MAMBO programme under grant agreement No.101060639, TETTRIs Grant Agreement 101081903, and Observation.org.
\bibliographystyle{plainnat}
 
\end{document}